# Sistema sensor para el monitoreo ambiental basado en redes Neuronales

## *Sensor System Based in Neural Networks for the Environmental Monitoring*

Rubio José de Jesús
*Instituto Politécnico Nacional
SEPI, ESIME Azcapotzalco
Correo: jrubioa@ipn.mx*

Hernández-Aguilar José Alberto
*Universidad Autónoma del Estado de Morelos
Correo: Jose_hernandez@uaem.mx*

Ávila-Camacho Francisco Jacob
*Tecnológico de Estudios Superiores de Ecatepec
Correo: favila11@udavinci.edu.mx*

Stein-Carrillo Juan Manuel
*Tecnológico de Estudios Superiores de Ecatepec
Correo: jstein11@udavinci.edu.mx*

Meléndez-Ramírez Adolfo
*Tecnológico de Estudios Superiores de Ecatepec
Correo: amelendez11@udavinci.edu.mx*

## Resumen

En las tareas de monitoreo ambiental resulta de gran importancia contar con sistemas compactos y portátiles capaces de identificar contaminantes ambientales que faciliten las tareas relacionadas con el manejo de los residuos y la restauración ambiental. En este trabajo se describe el desarrollo de un sistema sensor prototipo creado para identificar contaminantes en el ambiente. Este prototipo está conformado con un arreglo de sensores de gas de óxido de estaño $SnO_2$ utilizados para identificar vapores químicos, una etapa de adquisición de datos implementada con una plataforma ARM (Advanced RISC Machine) de bajo costo (Arduino) y una red neuronal capaz de identificar contaminantes ambientales automáticamente. La red neuronal se utiliza para identificar la composición del contaminante censado. En el sistema de cómputo, la carga computacional intensa se presenta únicamente en el proceso de entrenamiento, una vez que la red neuronal es entrenada, la operación consiste en propagar los datos a través de la red con una carga computacional mucho más ligera, la cual consiste principalmente en una multiplicación vector-matriz y una búsqueda en tablas que lleva a cabo la función de activación para identificar rápidamente muestras desconocidas.

**Descriptores:**

- redes neuronales
- inteligencia artificial
- contaminación ambiental
- sensores
- reconocimiento de patrones

## Abstract

*In the tasks of environmental monitoring is of great importance to have compact and portable systems able to identify environmental contaminants that facilitate tasks related to waste management and environmental restoration. In this paper, a prototype sensor is described to identify contaminants in the environment. This prototype is made with an array of tin oxide SnO2 gas sensors used to identify chemical vapors, a step of data acquisition implemented with ARM (Advanced RISC Machine) low-cost platform (Arduino) and a neural network able to identify environmental contaminants automatically. The neural network is used to identify the composition of contaminant census. In the computer system, the heavy computational load is presented only in the training process, once the neural network has been trained, the operation is to spread the data across the network with a much lighter computational load, which consists mainly of a vector-matrix multiplication and a search table that holds the activation function to quickly identify unknown samples.*

**Keywords:**

- *neural networks*
- *artificial intelligence*
- *environmental pollution*
- *sensors*
- *pattern recognition*

## Introducción

Las grandes cantidades de desperdicios y desechos que se generan todos los días en las ciudades producen desequilibrios ambientales que afectan la salud de sus habitantes. La generación más alta de residuos proviene de los hogares, se estima que estos generan 47% del total de los desperdicios, 29% los comercios y el resto se genera de otras actividades. Muchos de los desperdicios que se generan en cada casa habitación podrían no ser basura y convertirse en residuos aprovechables (Herrera, 2004).

Uno de los problemas que se generan con los residuos es la contaminación tanto de la tierra como del aire y el agua. La contaminación del aire se lleva a cabo por la descomposición de la materia orgánica, incendios y por los residuos y bacterias que se dispersan por el viento, el agua superficial se contamina por la basura que se arroja en ríos y cañadas, sin embargo, en los lugares donde se concentra la basura se filtran líquidos conocidos como lixiviados, que contaminan el agua del subsuelo y los mantos acuíferos. En el campo, la basura cambia la composición química del suelo y obstruye la germinación y el crecimiento de la vegetación (Tsai, 2008).

Durante el proceso de descomposición de los residuos orgánicos se desarrollan numerosos mohos y otros microorganismos que generan la emisión de esporas que representan un peligro potencial para el medio ambiente. La intensidad de la contaminación del aire a través de las emisiones de esporas de hongos de moho se correlaciona estrechamente con las condiciones climáticas, como las variaciones de temperatura ambiental, humedad relativa y radiación solar, entre otras afectaciones (Weinrich *et al.*, 1999).

Las tareas de monitoreo ambiental abarcan una amplia gama de actividades, ya que la contaminación del medio ambiente ocurre no solo por la descarga de desperdicios en el agua, la tierra y el aire, sino también por la generación de ruidos en el audio y en el rango de las frecuencias de telecomunicaciones. En los últimos años, se han desarrollado diversos sistemas sensores que cubren todas estas aplicaciones (Pearce *et al.*, 2003).

Este trabajo se centra en el uso de la tecnología de narices electrónicas para monitorear los compuestos orgánicos volátiles presentes en el aire que se liberan cuando los residuos se vierten en el agua, la tierra o el aire. Como contribución principal de este trabajo se construyó un prototipo de nariz electrónica a bajo costo que permite detectar y clasificar los gases contaminantes en el aire. Las pruebas se llevaron a cabo utilizando 5 elementos presentes en el hogar, lo que demuestra una solución de rápida respuesta que puede desarrollarse como una solución portable.

Como parte de esta misión, se pretende explorar tecnologías que contribuyan a la restauración ambiental y al manejo de los residuos a costos aceptables y reducidos. Estos esfuerzos incluyen el desarrollo de sistemas portátiles y económicos capaces de identificar contaminantes en tiempo real. El objetivo de esta investigación es demostrar las posibles capacidades de procesamiento de información del paradigma de redes neuronales en el análisis de sensores. Como parte inicial de esta propuesta se involucra el desarrollo de un prototipo, el cual combina un arreglo de sensores con una red neuronal.

Este trabajo se organiza en cuatro secciones: la sección 1 discute el problema del análisis de datos multivariable para la clasificación de olores, la sección 2 analiza el uso de redes neuronales para la clasificación de olores, la sección 3 describe el desarrollo experimental realizado con el prototipo construido y el censado de olores, la sección 4 presenta los resultados obtenidos y su comparación con otros resultados presentados en otros proyectos similares, finalmente se presentan nuestras conclusiones y trabajos futuros.

## Análisis de datos multivariable para la clasificación de datos

En el análisis de datos así como en la máquina de aprendizaje y la quimiometría se han empleado diversas técnicas de reconocimiento de patrones, sin embargo, para seleccionar el algoritmo apropiado para aplicaciones con narices electrónicas, es importante entender la naturaleza fundamental de los datos que se analizarán (Pearce *et al.*, 2003). Para analizar los datos de una nariz electrónica se requiere comprender la relación entre el conjunto de variables independientes (las salidas del arreglo de $n$ sensores) con el conjunto de variables dependientes (clases de olores o concentraciones de componentes) utilizando el análisis multivariable (Agatonovic y Beresford, 2000).

El problema del análisis de datos multivariable en la clasificación de olores es un análisis cualitativo en cuanto a los patrones de los olores producidos por este tipo de instrumentos y se considera cuantitativo cuando se requiere calcular la concentración de los componentes individuales. Por ello, los algoritmos de reconocimiento de patrones y procesamiento de datos son un componente crítico en la exitosa implementación y desarrollo de narices electrónicas (Pearce *et al.*, 2003).

Los algoritmos de reconocimiento de patrones comúnmente se clasifican en términos de ser paramétricos o no paramétricos y supervisados o no supervisados (Hines *et al.*, 2003):

- *Paramétricos*. Las técnicas paramétricas son aproximaciones estadísticas basadas en la suposición de que el espectro de los datos censados puede describirse por una función de densidad probabilística. En la mayoría de los casos, la suposición se basa en considerar que el flujo de datos constituye una distribución normal con una media y una varianza constantes. Estas técnicas intentan encontrar una relación, matemáticamente formulada, entre las entradas y las salidas.
- *No paramétricos*. Los métodos no paramétricos no realizan una suposición sobre una función de densidad probabilística y, por lo tanto, se aplican de forma general, ya que este tipo de técnicas multiva-riables para el análisis de datos se encuentran en los campos de las redes neuronales artificiales y los sistemas expertos.

- *Supervisados*. En un método de reconocimiento de patrones con aprendizaje supervisado, un conjunto de olores conocidos se introduce en la nariz electrónica de manera sistemática, donde los clasifica de acuerdo con descriptores o clases conocidas, agregándose a la base de conocimiento, posteriormente un olor desconocido se prueba contra la base de conocimiento para predecir la clase. El olor desconocido se analiza utilizando la relación creada en el entrenamiento dentro de un conjunto de olores conocidos.
- *No supervisados*. Los métodos con aprendizaje no supervisado aprenden separando las diferentes clases de manera rutinaria, discriminando entre los olores desconocidos. Este método utiliza un esquema de asociaciones intuitivas sin conocimiento previo, lo que los hace más cercanos al sistema olfativo humano.

## Uso de las redes neuronales para la clasificación de olores

Las redes neuronales artificiales se utilizan en una gran variedad de aplicaciones de procesamiento de datos, donde se requiere la extracción de información y el análisis de los datos en tiempo real. Una de las principales ventajas de las redes neuronales es que la mayor carga computacional se requiere únicamente en la etapa de entrenamiento. Una vez que la red neuronal se entrena para una tarea en particular, su operación es relativamente rápida y con ello se pueden identificar rápidamente muestras desconocidas (Ludermir y Yamazaki, 2003).

## Censado de olores mediante sistema prototipo

### Análisis de los datos censados

Los sensores recolectan la información del medio ambiente devolviendo valores que representan el compuesto químico censado y un valor relacionado con la concentración presente de dicho compuesto.

La figura 1 muestra el esquema de un sistema de monitoreo ambiental genérico, que se construye con la combinación de un arreglo de sensores y una red neuronal.

La cantidad y complejidad de los datos recolectados por los sensores representan una dificultad para el análisis convencional de los datos, pero con las redes neu-ronales utilizadas para el análisis complejo de los mismos y el reconocimiento de patrones, este análisis se facilita convirtiéndose en una alternativa ideal para el análisis de datos de sensores. El módulo de censado se construye con un arreglo de sensores, donde cada sensor en el arreglo se configura para responder a un componente específico. De esta forma, el número de sensores tendría que ser igual o mayor al número de compuestos que se pretende monitorear. Cuando se combina con una red neuronal, el número de compuestos detectables es mayor al número de sensores (Berna, 2010).

El arreglo de sensores se conforma de varios elementos sensores, donde cada elemento mide una propiedad diferente de la muestra censada. Cada químico en estado gaseoso presente en el arreglo de sensores produce una marca o patrón característico del gas en cuestión. Al presentar varios compuestos químicos en el arreglo de sensores, se construye una base de datos de los patrones o marcas generados con el arreglo de sensores. Esta base de datos con las marcas etiquetadas se utiliza para entrenar al sistema de reconocimiento de patrones. El objetivo de este proceso de entrenamiento es configurar el sistema de reconocimiento para producir una clasificación única para cada químico y con ello implementar el sistema de identificación automatizado (Keller, 1995).

El conjunto de datos de entrenamiento se utiliza para configurar la red neuronal, con el objetivo de aprender una asociación existente entre los patrones del arreglo de sensores y las etiquetas que representan los datos. Esta combinación entre sistemas sensores de gas y redes neuronales para identificar vapores, olores y aromas se conoce como nariz electrónica o nariz artificial (Kubiak, 2003).

Diversos autores han desarrollado narices electrónicas que se incorporan con redes neuronales para aplicaciones que involucran el monitoreo de olores en alimentos y bebidas, el control automatizado de sabores, el análisis de mezclas de combustibles, así como cuantificar componentes individuales en gases (Schaller *et al.*, 1998).

En aplicaciones de narices electrónicas se han incorporado diversas configuraciones de redes neuronales entre las cuales se incluyen el entrenamiento de propagación hacia atrás, redes de alimentación hacia adelante, redes auto organizadas de Kohonen, redes Hamming, máquinas Boltzman y redes Hopfield principalmente (Enedettia *et al.*, 2004).

## Etapa de censado de vapores químicos

El prototipo propuesto en este estudio se muestra en las figuras 2, 3 y 4, el prototipo tiene el objetivo de identificar y cuantificar vapores químicos. El prototipo emplea un arreglo de 5 sensores de óxido de estaño, un sensor de humedad y un sensor de temperatura para examinar el ambiente.

Cada sensor se diseñó para un químico en específico y cada uno responde a una variedad amplia de vapores químicos. En forma colectiva, los sensores responden con una marca única (patrón) a diferentes químicos. Durante el proceso de entrenamiento, se exponen varios químicos al arreglo de sensores y con ello se obtienen una serie de valores en un lapso de tiempo. Se utilizó el algoritmo de propagación hacia atrás para entrenar a la red neuronal y con ello proporcionar el análisis correcto para el químico presente.

Se utilizaron sensores de gas de la marca Figaro, que son productos comerciales (Sensor 1 MQ-2, Sensor 2 MQ-135, Sensor 3 MQ 3, Sensor 4 TGS 2610 y Sensor 5 TGS 2611). El sensor de humedad y temperatura (HM-Z433A1) se utilizó para monitorear las condiciones del experimento y también se envió a la red neuronal.

## Etapa de identificación y clasificación

Como se mencionó anteriormente, el problema de reconocimiento de patrones en los datos de las narices electrónicas está estrechamente ligado al análisis de datos multivariables, la figura 5 resume las principales técnicas de procesamiento multivariable que se han utilizado en aplicaciones de narices electrónicas.

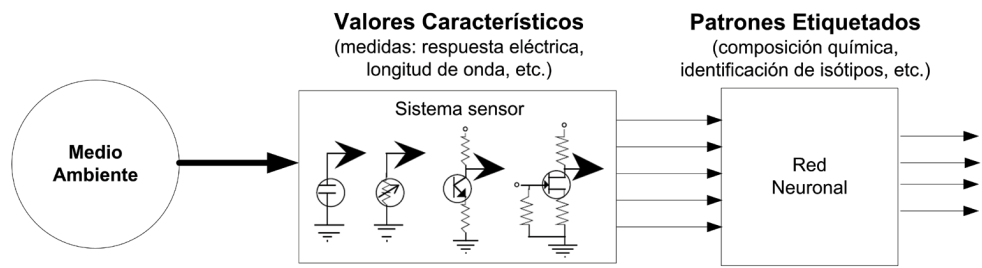

Figura 1. Sistema de censado combinado con una red neuronal.

Fuente: Keller *et al.* (1994)

El esquema de clasificación se lleva a cabo en tres niveles, primero se hace una distinción entre aproximaciones estadísticas y biológicas, después entre el análisis de patrones cualitativos y cuantitativos, para finalmente hacer una distinción entre técnicas supervisadas y no supervisadas (Bucak y Karlık, 2009).

La *función de análisis discriminante* (DFA) es un clasificador paramétrico de aprendizaje supervisado, que puede utilizarse para el análisis tanto cuantitativo como cualitativo. El *análisis de componente principal* (PCA), es un método de proyección no paramétrico comúnmente utilizado para implementar un clasificador supervisado lineal, en conjunto con el análisis discriminante (Hines *et al.*, 2003).

Asimismo, muchos de los procedimientos aceptados que se utilizan en el reconocimiento de patrones tradicional, no siempre son pertinentes o relevantes cuando se aplican al reconocimiento de patrones en narices electrónicas. En muchos casos, se espera que las narices electrónicas operen en diferentes ambientes y situaciones y el esquema de reconocimiento sea viable con estas situaciones, como en los casos donde se requiere realizar las mediciones en campo, teniendo retos adicionales a los que se tienen en laboratorio o en ambientes controlados, en campo se espera que los sistemas detecten e identifiquen los elementos que se analicen aún cuando existan componentes desconocidos que interfieran. Por lo que existen algunos criterios o cualidades ideales que se espera cumplan los algoritmos de reconocimiento de patrones tales como, precisión velocidad o habilidad para hacer frente a la incertidumbre (Hines *et al.*, 2003).

Para muchos investigadores que trabajan en los campos de los sistemas de narices electrónicas las dos principales técnicas de reconocimiento de patrones son PCA (*análisis de componente principal*) para desplegar olores conocidos, explorar el agrupamiento de los datos en el espacio multisensor y evaluar su separabilidad lineal, así como las redes neuronales con algoritmo de entrenamiento de propagación hacia atrás, con el fin de proporcionar una clasificación predictiva de olores desconocidos. Sin embargo PCA solo

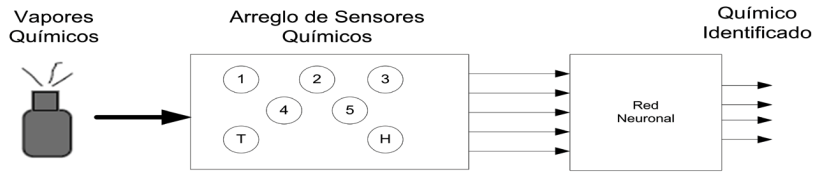

Figura 2. Etapa de censado de vapores químicos

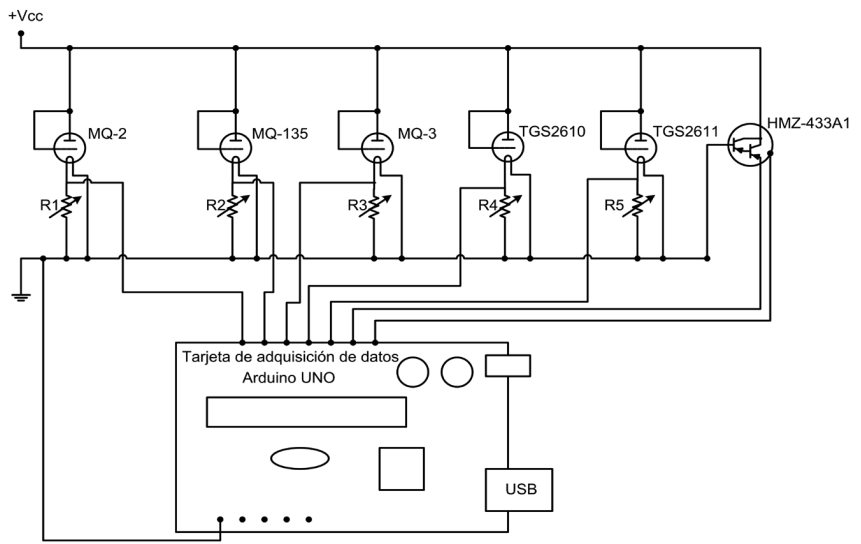

Figura 3. Diagrama electrónico del arreglo de sensores químicos

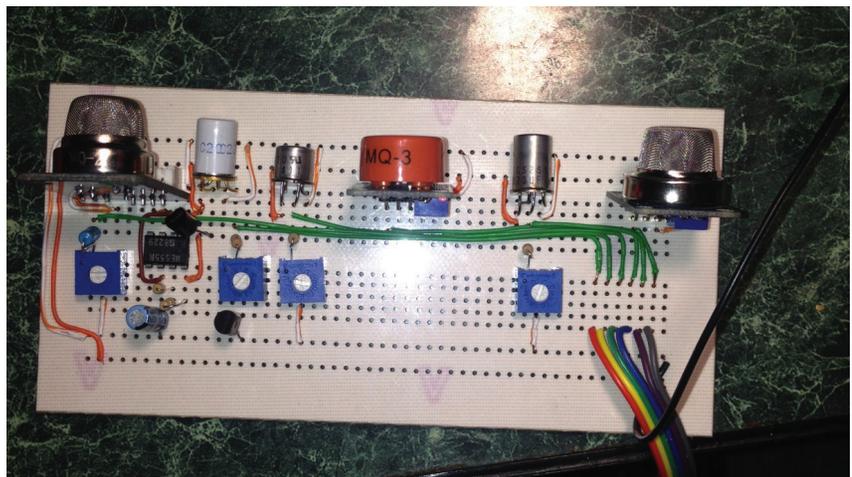

Figura 4. Prototipo del arreglo de sensores desarrollado para la detección de gases químicos

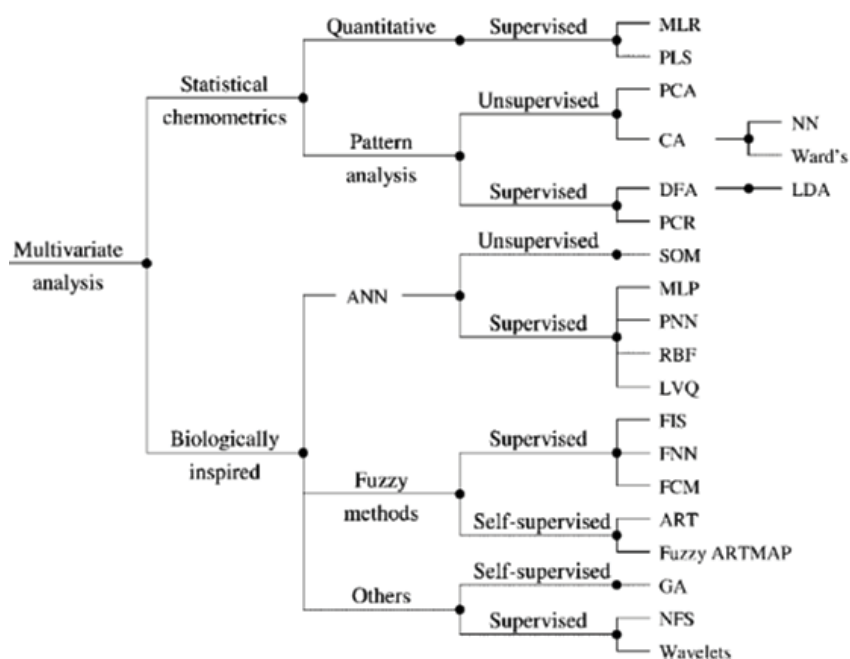

Figura 5. Esquema de técnicas de procesamiento multivariable utilizadas en conjuntos de datos proporcionados por narices electrónicas (Pearce *et al*., 2003)

puede utilizarse para proporcionar una representación lineal de las clases y no para propósitos de clasificación, por lo que las redes neuronales son las más exitosas en muchas aplicaciones que se enfocan en la discriminación de olores disimilares simples o complejos o el estancamiento de olores complejos específicos (Pearce *et al*., 2003).

Las distintas configuraciones de redes neuronales se han convertido en los clasificadores más precisos que pueden hacer frente a grupos, clases o clusters que se observa se sobreponen con técnicas lineales, razón por la cual las narices electrónicas comerciales hoy en día ofrecen un clasificado basado en redes neuronales con algoritmo de entrenamiento de propagación hacia atrás. De acuerdo con Pearce *et al*. (2003), la mejor estrategia para el reconocimiento de patrones en narices electrónicas es utilizar algoritmos, salvo algunas excepciones, que puedan afrontar un cierto grado de borrosidad como en el caso del sistema olfativo humano. En este sentido, una combinación entre redes neuronales y lógica difusa se presenta como una alternativa atractiva que podría ejecutar un aprendizaje incremental y un potencial auto organizado y auto estabilizado. Por lo que este tipo de soluciones serán más utilizadas en un futuro para producir clasificaciones basadas en un conjunto de reglas entendibles (Pearce *et al*., 2003).

Analizando los datos presentados por los sensores para los 5 compuestos analizados, se observa que existen una frontera de decisión no lineal y que además se tienen espacios sobrepuestos entre los diferentes compuestos, por lo que no es factible utilizar algunos de los métodos estadísticos, ya que el problema es la clasificación de 5 tipos de olores, más que la determinación de la concentración de cada uno de ellos, el análisis discriminante o el análisis de cluster no facilitarían la clasificación de los compuestos analizados, la figura 6 muestra una gráfica de dispersión para los datos tanto de gas líquido para encendedores como para el alcohol, donde se observa que los espacios se interponen con los diferentes sensores.

En este sentido, se decidió utilizar en una primera solución, redes neuronales como el sistema de reconocimiento de patrones aplicado al prototipo de nariz electrónica construido para el proyecto de investigación. Con ello se podrán establecer los argumentos para desarrollar sistemas más complejos basados en la lógica difusa como siguiente aproximación. Para ello se decidió establecer la mejor configuración de una red neuronal de tres capas comparando dos configuraciones en las que la diferencia principal radica en la cantidad de neuronas de la capa oculta para con ello determinar la complejidad de la red necesaria para la clasificación de los olores de cinco compuestos.

El prototipo de la red neuronal se creó utilizando toolbox de Matlab, así como para la elaboración de pruebas comparando los resultados de dos tipos de redes neuronales multicapa de alimentación hacia delante, donde fueron entrenadas con el algoritmo de propagación hacia atrás utilizando un conjunto de entrenamiento tomado con las lecturas de 4 compuestos.

Los parámetros utilizados para entrenar a las dos redes neuronales se muestran en la tabla 1, donde se describe que el algoritmo de entrenamiento es de propagación hacia atrás (*backpropagation*), 7 neuronas en la capa de entrada, una para cada uno de los 5 sensores de gas, una para el sensor de temperatura y una más para el sensor de humedad. La primera red neuronal tiene 3 neuronas en la capa oculta y la segunda tiene 10 neuronas en dicha capa, finalmente en la etapa de salida ambas redes se conformaron con 5 neuronas en la capa de salida.

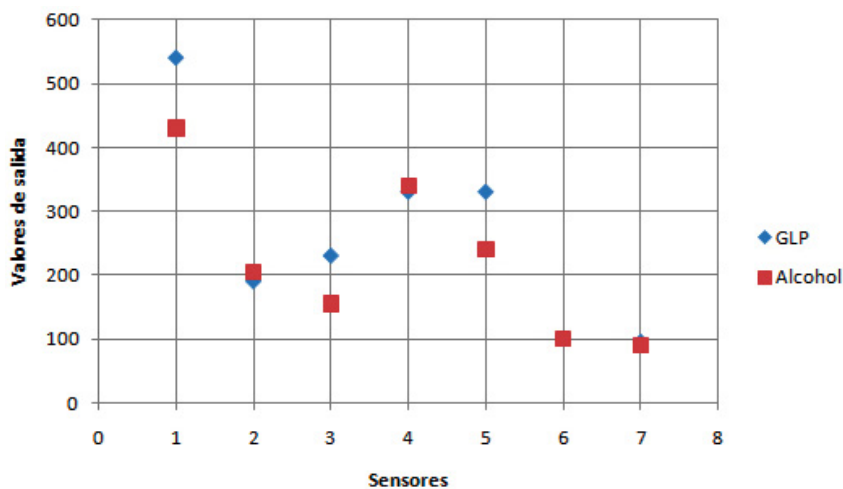

Figura 6. Gráfica de dispersión para los valores censados del gas líquido (GLP) y el alcohol

Tabla 1. Parámetros de entrenamiento de las redes neuronales

| Características | Red 1 | Red 2 |
|---|---|---|
| Tipo | Propagación hacia atrás | Propagación hacia atrás |
| Arquitectura | 7 – 3 – 5 Feedforward | 7 – 10 – 5 Feedforward |
| Función de activación | Log-Sigmoid | Log-Sigmoid |
| Taza de aprendizaje | 0.01 | 0.01 |
| Momentum | 0.9 | 0.9 |
| Núm. de Epochs | 1000 | 1000 |

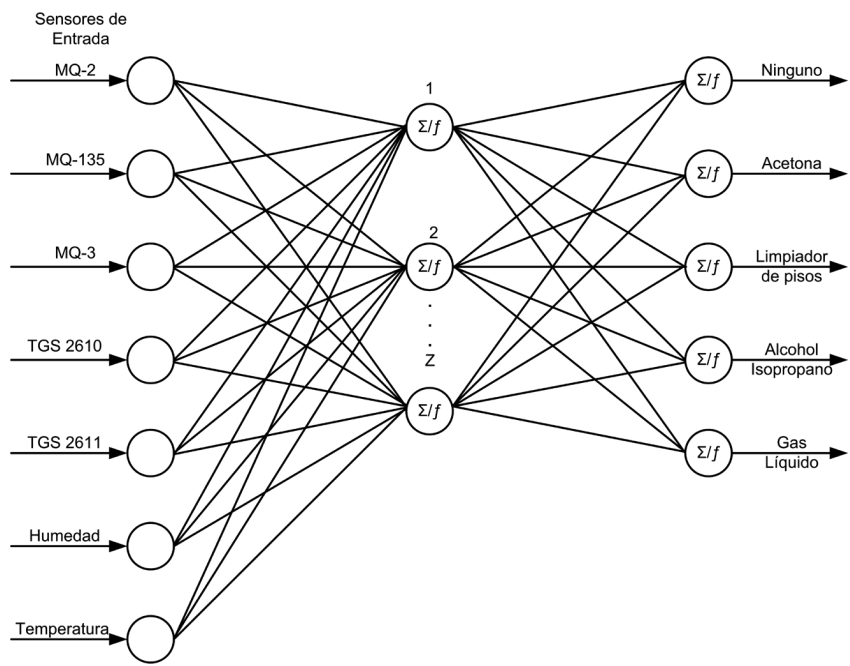

Figura 7. Arquitectura de las redes neuronales utilizadas para identificar químicos presentes en el hogar. Para la Red 1 Z=3, 3 neuronas en la capa oculta y para la red 2 Z = 10, con 10 neuronas en la capa oculta

El prototipo se entrenó inicialmente para identificar químicos comunes en el hogar: acetona, limpiador de pisos, alcohol isopropano y gas líquido para encendedores. Se agregó una categoría más (ninguno) para denotar la ausencia de todos ellos, excepto aquellos normalmente presentes en el aire. Esto conformó una salida de la red neuronal de 5 categorías, generando una salida de 5 neuronas en la capa final. En cada neurona de la capa oculta y la salida, la función de activación y transferencia es la función *log-sigmoid* para ambas redes neuronales.

La figura 7 muestra las arquitecturas de las redes neuronales, para la red neuronal 1 Z = 3 y para la red neuronal 2, Z = 10, utilizadas para identificar los compuestos químicos.

En la figura 8 se muestran las redes neuronales creadas con toolbox.

Durante la operación, el arreglo de sensores detecta olores, las señales de los sensores se digitalizan, se envían a la computadora y la red neuronal implementada con toolbox identifica los químicos. El tiempo de identificación está limitado únicamente por el tiempo de respuesta de los sensores químicos, pero el proceso completo se lleva a cabo en unos cuantos segundos.

Determinar cuál de las dos redes neuronales presenta el mejor resultado para la identificación de los compuestos químicos, está en función del porcentaje de aciertos y errores presentados en cada una de las pruebas.

**Resultados**

Las figuras 9 y 10 muestran la respuesta de los sensores y la clasificación que se llevó a cabo por cada una de las redes neuronales en pruebas realizadas con los químicos presentes en el prototipo.

Las figuras también muestran dos gráficas para cada compuesto

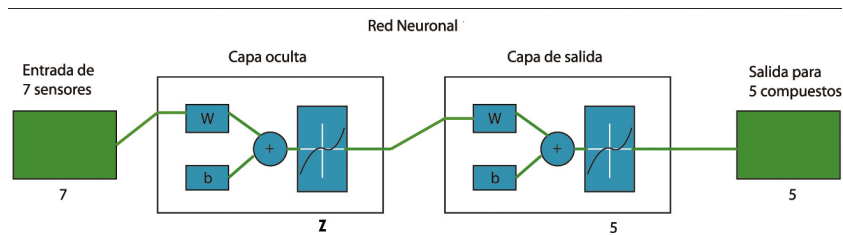

Figura 8. Estructura de las redes neuronales 1 y 2 creadas en toolbox para identificar químicos presentes en el hogar, para la red 1 Z = 3 y para la red 2 Z = 10

químico censado, la gráfica del lado izquierdo muestra los valores de respuesta de cada sensor de gas para el compuesto químico correspondiente, donde los valores del eje x corresponden a los sensores de la siguiente forma: 1 = MQ-2, 2 = MQ-135, 3 = MQ-3, 4 = TGS2610, 5 = TGS2611, 6 = Sensor de humedad y 7 = Sensor de temperatura. Los valores del eje y son los valores de respuesta de cada sensor en respuesta al censado del compuesto químico.

Las gráficas del lado derecho, en las mismas figuras, representan la salida de la red neuronal correspondiente para cada uno de los compuestos censados, donde el valor 1 indica la presencia del compuesto. Los valores en el eje x representan cada una de las salidas que se corresponde con el compuesto químico correspondiente, de la siguiente forma:

1 = S1 (Ningún compuesto presente), 2 = S2 (Acetona), 3 = S3 (Limpiador de pisos), 4 = S4 (Alcohol isopropano) y 5 = S5 (Gas líquido para encendedores).

De esta forma, cuando en el sistema se tiene la presencia de Acetona, los valores en las 5 salidas de la red neuronal deberán ser: 0,1,0,0,0.

En las gráficas se observa que algunos valores en las salidas que representan compuestos químicos no presentes en ambas redes neuronales son mayores a 0, lo que muestra falsos positivos en la respuesta de las redes neuronales. La configuración más adecuada para la solución es la que presenta la menor tasa de falsos positivos.

Los sistemas de detección de olores aplican en diversas ejecuciones industriales, incluyendo el monitoreo de la calidad del aire en hogares (Keller *et al.*, 1994; Ras *et al.*, 2010), el cuidado de la salud, la seguridad (Casalinuovo *et al.*, 2006), el monitoreo ambiental, la calidad de los productos alimenticios (Berna, 2010), (Dębska y Guzowska, 2011), (Di Natale *et al.*, 2000), (Kubiak, 2003), diagnósticos médicos (Keller, 1995; Wilson y Baietto, 2011), identificación de personas (Wongchoosuk *et al.*, 2009; Porras y Salinas, 2011; Stitzel *et al.*, 2011), así como en aplicaciones farmacéuticas y militares para la detección de gases peligrosos, por mencionar algunas (Wilson y Baietto, 2011; Wilson, 2013).

La investigación en los métodos alternativos para la detección de olores logró avances importantes desde sus inicios en 1982 (Persaud y Dodd, 1982). Durante los últimos años existió una gran variedad de desarrollos para construir instrumentos que funcionen como narices electrónicas (Wilson y Baietto, 2009). Existen algunos productos comerciales, pero en su mayoría son equipos voluminosos y de precios altos, que hacen complicada una adopción generalizada de los mismos. Muy pocos productos son portátiles, pero aun así requiere de equipos especializados y muchos de ellos están enfocados a tareas particulares de áreas específicas dados los tipos de sensores utilizados en cada aplicación (Tang *et al.*, 2010), razón por la cual muchos de esos sistemas no son factibles de comercializar como productos portables, sino más bien se utilizan como equipos de laboratorio.

Por otro lado, en Gómez *et al.* (2007; 2006) los autores utilizan una nariz electrónica para monitorear la madurez del tomate durante su vida en estantería. Para ellos la tecnología de la nariz electrónica ofrece una alternativa no destructiva y eficaz para medir el aroma y así obtener la información para acceder al estado de maduración de frutas y legumbres durante su vida en estantería. Su objetivo fue evaluar la capacidad de la nariz electrónica para monitorear los cambios en la producción de compuestos volátiles del tomate durante dos tratamientos de almacenamientos diferentes. Los autores utilizaron el análisis de componente principal PCA y el análisis del discriminante lineal LDA para distinguir los tomates con diferentes tiempos de almacenamiento.

Los resultados prueban que la nariz electrónica utilizada puede diferenciar satisfactoriamente los estados de maduración del tomate a través de su tiempo de almacenamiento. Los análisis PCA y LDA clasificaron 97.77 y 95.55%, respectivamente, del total de muestras (90 muestras) en sus respectivos grupos (5 grupos). La nariz electrónica empleada tiene un arreglo de 10 sensores MOS.

Con una cantidad menor de sensores utilizando redes neuronales en lugar de métodos estadísticos, los resultados obtenidos con el prototipo construido obtuvieron 100% de eficiencia en la clasificación de los olores.

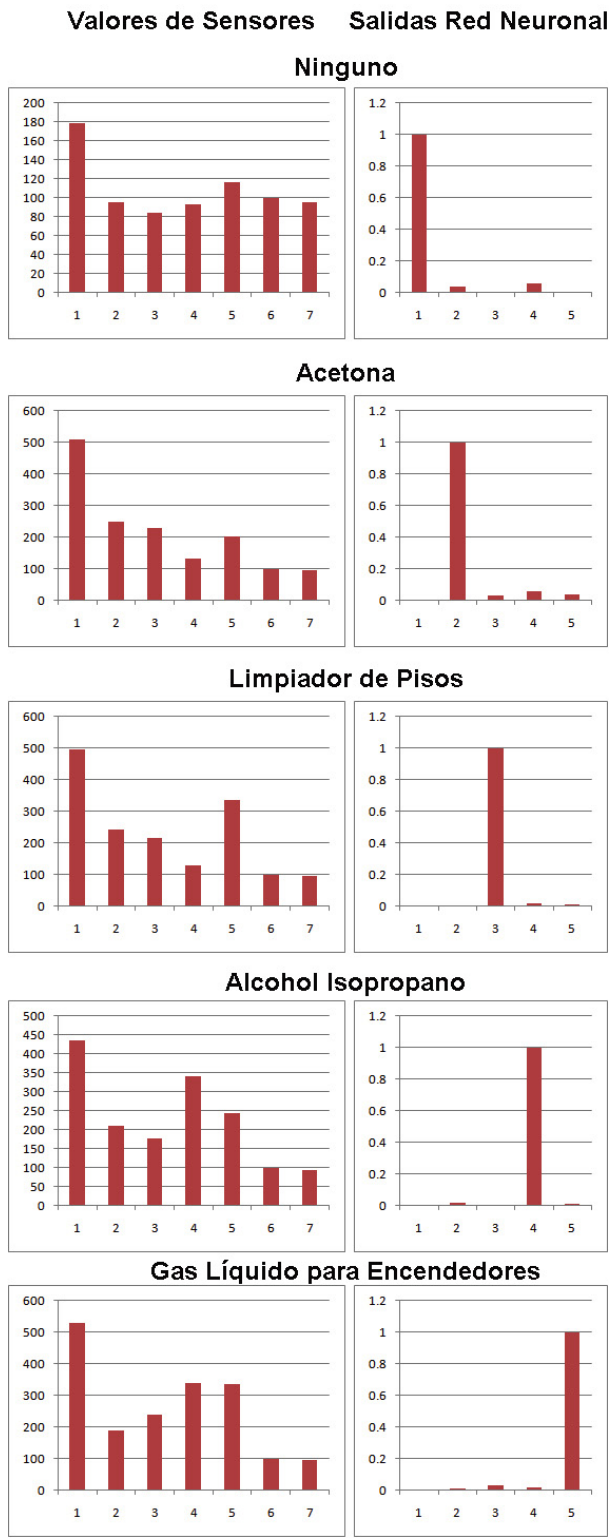

Figura 9. Respuestas de los sensores a las muestras y clasificación de la red neuronal 1. Los números en el eje x corresponden a los sensores en las gráficas del lado izquierdo y a las salidas de la red neuronal en las gráficas del lado derecho

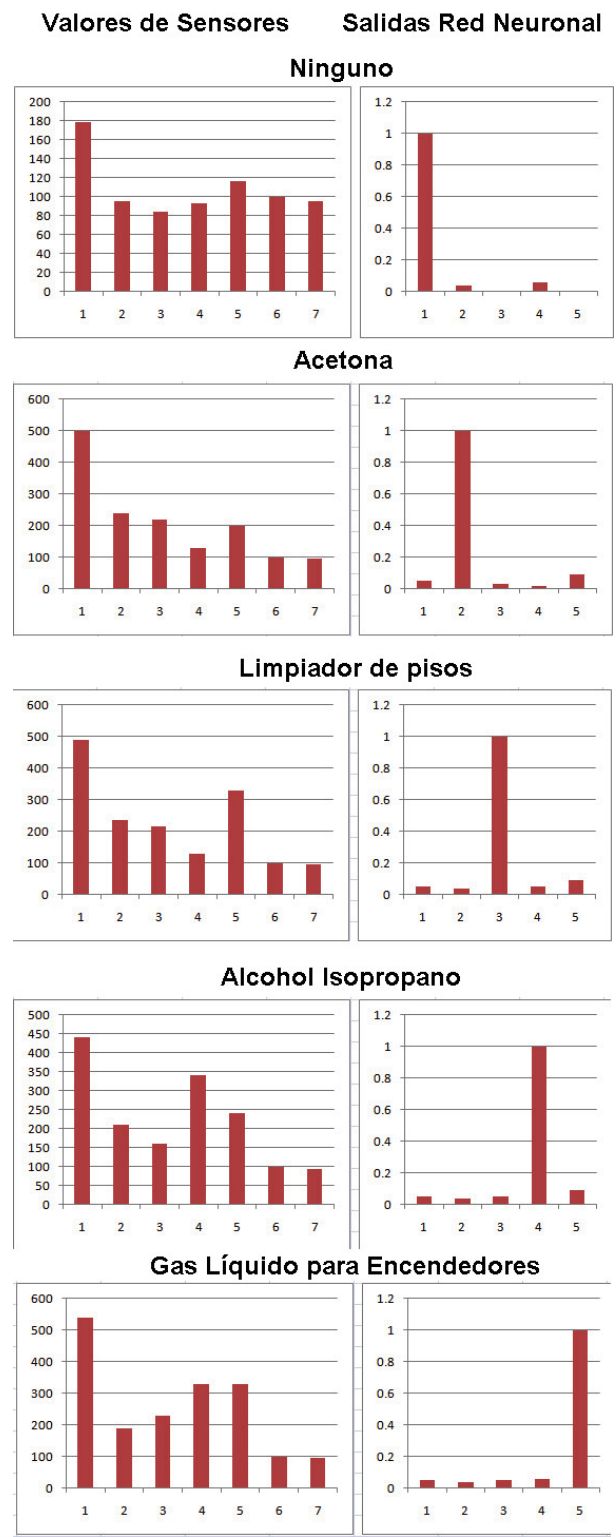

Figura 10. Respuestas de los sensores a las muestras y clasificación de la red neuronal 2. Los números en el eje x corresponden a los sensores en las gráficas del lado izquierdo y a las salidas de la red neuronal en las gráficas del lado derecho

## Conclusiones y trabajos futuros

El prototipo desarrollado combina un arreglo de sensores de gas de óxido de estaño con una red neuronal utilizada para identificar químicos comunes presentes en el hogar. Los resultados iniciales demuestran la capacidad para el reconocimiento de patrones del paradigma de las redes neuronales en el análisis de sensores. Al mismo tiempo, el prototipo es un desarrollo compacto y portable que facilita el análisis en tiempo real, así como la automatización del proceso de censado, análisis y reconocimiento.

En cuanto a las dos redes neuronales creadas, la primera presenta una tasa de falsos positivos menor a la siguiente, aunque en ambas, dicha tasa de falsos positivos es pequeña y aceptable para el desarrollo de soluciones, sin embargo la primera, que únicamente contiene 3 neuronas en la capa oculta, presenta una tasa de falsos positivos menor y al mismo tiempo una carga de procesamiento mucho menor que la red con 10 neuronas en la capa oculta. Con este hecho resulta aceptable incorporar algoritmos de redes neuronales en sistemas portables.

Para trabajos futuros se propone comparar las redes neuronales para el análisis de sensores con las técnicas convencionales, explorar otro tipo de paradigmas de redes neuronales e involucrar los prototipos con los sistemas de campo.

Al mismo tiempo se concluye que el monitoreo ambiental es un área prometedora para aplicaciones de la tecnología de narices electrónicas.

## Agradecimientos

## Referencias

Agatonovic-Kustrin S. y Beresford R. Basic concepts of artificial neural network (ANN) modeling and its application in pharmaceutical research. *Journal of Pharmaceutical and Biomedical Analysis*, volumen 22 (número 5), 2000: 717-727, doi:10.1016/S0731-7085(99)00272-1

Berna A. Metal oxide sensors for electronic noses and their application to food analysis. *Sensors*, 2010, doi:10.3390/s100403882

Bucak İ.Ö. y Karlık B. Hazardous odor recognition by CMAC based neural networks. *Sensors*, 2009, doi:10.3390/s90907308.

Casalinuovo I.A., Di Pierro D., Coletta M., Di Francesco P. Application of electronic noses for disease diagnosis and food spoilage detection. *Sensors*, 2006, doi:10.3390/s6111428

Dębska B. y Guzowska-Świder B. Decision trees in selection of featured determined food quality. *Analytica Chimica Acta*, volumen 705 (números 1-2), 2011: 261-71, doi:10.1016/j.aca.2011.06.030

Di Natale C., Paolesse R., Macagnano A., Mantini A., D'Amico A., Legin A., Vlasov Y. Electronic nose and electronic tongue integration for improved classification of clinical and food samples. *Sensors and Actuators B: Chemical*, 2000, doi:10.1016/S0925-4005(99)00477-3

Enedettia S.B., Anninoa S.M., Abatinib A.G.S., Luigi G. Electronic nose and neural network use for the classification of honey, volumen 35, 2004:1-6, doi:10.1051/apido

Gómez A.H., Pereira A.G., Wang J. Using electronic nose technique to monitoring tomato maturity states during shelf live Uso de la técnica de la nariz electrónica para monitorear la madurez del tomate durante su vida en estantería, volumen 16, 2007: 24-30.

Gomez A., Hu G., Wang J., Pereira A. Evaluation of tomato maturity by electronic nose. *Computers and Electronics in Agriculture*, volumen 54 (número 1), 2006: 44-52, doi:10.1016/j.compag.2006.07.002

Herrera-Massieu R. *Ley de residuos sólidos para el Distrito Federal*, México, DF, 2004.

Hines E.L., Boilot P., Gardner J.W., Gongora M. Pattern analysis for electronic noses. *Handbook of Machine Olfactory*, volumen 1, 2003: 645.

Keller P.E. Electronic noses and their applications, IEEE Technical Applications Conference and Workshops Northcon95 Conference Record, 1995, doi:10.1109/NORTHC.1995. 485024

Keller P.E., Kouzes R.T., Kangas L.J., Box P.O. Three neural network based sensor systems for environmental monitoring, 1994, pp. 378-382.

Kubiak A. Artificial nose and its application to food evaluation. *ProblemyInzynieriiRolniczej*, volumen 11 (número 2), 2003: 27-34.

Ludermir T.B. y Yamazaki A. Neural networks for odor recognition in artificial noses. *Proceedings of the International Joint Conference on Neural Networks*, volumen 1, 2003,doi:10.1109/IJCNN.2003.1223317

Pearce T.C., Schiffman S.S., Nagle H.T., Gardner J.W. *Handbook of machine olfaction: electronic nose technology*. Weinheim, 3 WILEY-VCH Verlag GmbH & Co. KGaA, 2003.

Persaud K. y Dodd G. Analysis of discrimination mechanisms in the mammalian olfactory system using a model nose. *Nature*, volumen 299 (número 5881), 1982: 352-355.

Porras-Chavarino C. y Salinas-Martínez de Lecea J.M. Artificial neural networks in Neurosciences. *Psicothema*, volumen 23 (número 4), 2011:738-744, doi:10.1016/S1383-7621(97)00063-5

Ras M.R., Marcé R.M., Borrull F. Volatile organic compounds in air at urban and industrial areas in the Tarragona region by thermal desorption and gas chromatography-mass spectrometry. *Environmental Monitoring and Assessment*, volumen 161, (números 1-4), 2010: 389-402, [en linea]. Disponible en: http://www.ncbi.nlm.nih.gov/pubmed/19238572

Schaller E., Bosset J.O., Escher F. Electronic noses and their application to food. *LWT - Food Science and Technology*, volumen 31 (número 4), 1998: 305-316, doi:10.1006/fstl.1998.0376

Stitzel S.E., Aernecke M.J., Walt D.R. Artificial noses. *Annual Review of Biomedical Engineering*, volumen 13, 2011: 1-25, doi:10.1146/annurev-bioeng-071910-124633

Tang K.T., Chiu S.W., Pan C.H., Hsieh H.Y., Liang Y.S., Liu S.C. Development of a portable electronic nose system for the detection and classification of fruity odors. *Sensors*, 2010, doi:10.3390/s101009179

Tsai W.T. Management considerations and environmental benefit analysis for turning food garbage into agricultural resources. *Bioresource Technology*, volumen 99 (número 13), 2008: 5309-5316 [en linea]. Disponible en: http://www.sciencedirect.com/science/article/B6V24-4RWJVTC-3/2/c74774de3de6f0db-3096c01595658c47

Weinrich M., Vissiennon T., Kliche R., Schumann M., Bergmann A. Nature and frequency of the existence of mold fungi in garbage cans for biological waste and the resultant airborne spore pollution. *Berliner Und Munchener Tierarztliche Wochenschrift*, volumen 112 (número 12), 1999: 454-458.

Wilson A.D. Diverse applications of electronic-nose technologies in agriculture and forestry. *Sensors*, volumen 13 (número 2), 2013: 2295-348, doi:10.3390/s130202295

Wilson A.D. y Baietto M. Applications and advances in electronic-nose technologies. *Sensors Peterboroug*, voumen 9 (número 7), 2009: 5099-5148, doi:10.3390/s90705099

Wilson A.D. y Baietto M. Advances in electronic-nose technologies developed for biomedical applications. *Sensors (Basel, Switzerland)*, volumen 11 (número 1), 2011: 1105-76, doi: 10.3390/s110101105

Wongchoosuk C., Lutz M., Kerdcharoen T. Detection and classification of human body odor using an electronic nose. *Sensors*, volumen 9 (número 9), 2009: 7234-7249, doi:10.3390/s90907234

## Semblanzas de los autores

*José de Jesús Rubio*: Obtuvo la licenciatura por la ESIME Zacatenco del Instituto Politécnico Nacional, México. Obtuvo el grado de maestría y de doctorado en control automático del CINVESTAV IPN. Fue profesor de tiempo completo en la Universidad Autónoma Metropolitana durante 2 años. Actualmente es profesor de tiempo completo de la Sección de Estudios de Posgrado e Investigación, ESIME Azcapotzalco del Instituto Politécnico Nacional. Ha publicado 74 artículos en revistas internacionales, 1 libro internacional, 8 capítulos en libros internacionales, y ha presentado 29 trabajos en congresos internacionales con 416 citas. Es miembro de los sistemas adativos difusos de la IEEE. Forma parte del consejo editorial de la revista Evolving Systems. Fue director de 3 estudiantes de posdoctorado, 5 de estudiantes de doctorado, 29 estudiantes de maestría, 4 estudiantes de especialidad y 17 estudiantes de licenciatura. Sus intereses de investigación se centran principalmente en el modelado dinámico, la pasividad, la evolución de los sistemas inteligentes, sistemas inteligentes estables, control inteligente, control no lineal, control adaptativo, control por modos deslizantes, control óptimo, sistemas neuronales difusas, filtro de Kalman, mínimos cuadrados, elipsoide acotado, sistemas retardados, detector de colisiones, generador de trayectoria, reconocimiento de patrones, de identificación, predicción, procesamiento de imágenes, pilas de combustible, robótica, mecatrónica, médicos, energía alternativa, automotrices, cuadrotors, procesamiento de señales, invernaderos, derivados del petróleo, incubadoras, almacenes, reactores químicos, mezcladores, y narices electrónicas.

*José Alberto Hernández-Aguilar*. Doctor en ingeniería y ciencias aplicadas de la Universidad Autónoma del Estado de Morelos (UAEM). Master of Business Administration (MBA) de la Universidad de las Américas (UDLA), A.C., graduado con mención honorífica. Es ingeniero en computación por la UNAM. Desde 2010, se desempeña como profesor investigador de tiempo completo en la Facultad de Contaduría, Administración e Informática de la UAEM. Sus áreas de interés son: bases de datos, inteligencia artificial, minería de datos, optimización y procesamiento en paralelo.

*Francisco Jacob Ávila-Camacho*. Ingeniero en electrónica y sistemas digitales egresado de la Universidad Autónoma Metropolitana Unidad Azcapotzalco, tiene el grado de maestría en ingeniería en sistemas computacionales por parte del Tecnológico de Estudios Superiores de Ecatepec, cuenta con una maestría en ciencias en administración de negocios de la Escuela Superior de Comercio y Administración del IPN, es candidato a doctor en sistemas computacionales por la Universidad Da Vinci, actualmente se desempeña como profesor investigador en el Tecnológico de Estudios Superiores de Ecatepec, ha impartido ponencias en congresos nacionales e internacionales y publicado en memorias de congresos, actualmente es responsable y líder de un proyecto de investigación y desarrollo tecnológico financiado por la DGEST, adicionalmente colabora en proyectos de investigación en las áreas de inteligencia artificial, reconocimiento de patrones, minería de datos y sistemas embebidos. Es socio fundador de la firma Ihualia Software, S.A. de C.V., empresa con registro RENIECYT (Registro Nacional de Instituciones y Empresas Científicas y Tecnológicas), donde participa en el desarrollo de proyectos de innovación y desarrollo tecnológico.

*Juan Manuel Stein-Carrillo*. Es ingeniero en sistemas computacionales, egresado del Tecnológico de Estudios Superiores de Ecatepec, estudió dos maestrías, la primera en ingeniería en sistemas computacionales, y la segunda en administración y desarrollo de negocios con doble titulación por parte de la universidad del valle de Toluca (México) y la Universidad Politécnica de Cataluña (España). Actualmente es candidato a doctor en sistemas computacionales por la Universidad DaVinci, es profesor de tiempo completo el Tecnológico de Estudios Superiores de Ecatepec. Desarrolló actualmente un proyecto financiado por la DGESTen el área de TICS y en específico con interfaces humano computadora y reconocimiento de patrones.

*Adolfo Melendez-Ramirez*. Es ingeniero en comunicaciones y electrónica, egresado de la ESIME Zacatenco IPN, estudió una maestría en administración y desarrollo de negocios con doble titulación por parte de la universidad del valle de Toluca (México) y la Universidad Politécnica de Cataluña (España). Actualmente es candidato a doctor en sistemas computacionales, profesor de tiempo completo en el Tecnológico de Estudios Superiores de Ecatepec, desarrolla proyectos tecnológicos dentro de la institución donde trabaja con interfaces humano computadora y reconocimiento de patrones.